\documentclass{article}

\usepackage{microtype}
\usepackage{graphicx}
\usepackage{caption} 
\usepackage{subcaption}
\usepackage{booktabs} 
\usepackage{CJKutf8}

\usepackage{hyperref}

\usepackage[accepted]{whi2020}

\icmltitlerunning{Sequential Explanations with Mental Model-Based Policies}

\begin{document}

\twocolumn[
\icmltitle{Sequential Explanations with Mental Model-Based Policies}




\begin{icmlauthorlist}
\icmlauthor{Arnold YS Yeung}{uoft,vec}
\icmlauthor{Shalmali Joshi}{vec}
\icmlauthor{Joseph Jay Williams}{uoft}
\icmlauthor{Frank Rudzicz}{uoft,vec}
\end{icmlauthorlist}

\icmlaffiliation{uoft}{Department of Computer Science, University of Toronto, Toronto, Canada}
\icmlaffiliation{vec}{Vector Institute for Artificial Intelligence, Toronto, Canada}

\icmlcorrespondingauthor{Arnold YS Yeung}{arnoldyeung@cs.toronto.edu}

\icmlkeywords{Machine Learning, ICML}

\vskip 0.3in
]



\printAffiliationsAndNotice{}  

\begin{abstract}
The \textit{act of explaining} across two parties is a feedback loop, where one provides information on what needs to be explained and the other provides an explanation relevant to this information. We apply a reinforcement learning framework which emulates this format by providing explanations based on the \textit{explainee}'s current mental model. We conduct novel online human experiments where explanations generated by various explanation methods are selected and presented to participants, using policies which observe participants' mental models, in order to optimize an interpretability proxy. Our results suggest that mental model-based policies (anchored in our proposed state representation) may increase interpretability over multiple sequential explanations, when compared to a random selection baseline. This work provides insight into how to select explanations which increase relevant information for users, and into conducting human-grounded experimentation to understand interpretability.
\end{abstract}

\section{Introduction}

As machine learning becomes more commonly used by individuals without technical expertise, explanations of the model's behavior are necessary to build trust and confidence \cite{adadi2018peeking, miller2017explainable, doran2017does, hoffman2018metrics}. Decision makers should be able to interpret how models behave and assess the \textit{reasoning} behind specific predictions: This allows identification of features that may have been overlooked by humans, and any errors behind the output of these models. By gaining further insight into the task and evaluating the operation of the models, users may apply them with greater understanding, guaranteeing safer use, and ultimately encouraging further effective adoption of these approaches.

In this work, we focus on explanation methods, which elucidate how a black-box model behaves \cite{adadi2018peeking}. One criticism of many explanation methods is that they are generally designed by developers and ML experts whose notions of interpretability do not align with the users \cite{miller2017explainable}. As a result, explanations provided to users (i.e., the \textit{explainee}) may be irrelevant or not convey useful information.

The \textit{act of explaining} is an interaction between the explainer and the explainee, often in a dialogue \cite{adadi2018peeking, miller2017explainable, madumal2018towards}. More precisely, the explainer presents an explanation to the explainee, and the explainee attempts to understand it. The explainee often provides feedback to the explainer as an indication of their understanding (e.g., asking further questions) and the explainer then provides a following explanation based on the received feedback. We refer to any observations of this indication (i.e., a representation of the explainee's interpretation of the black-box model) as the explainee's \textit{mental model}. The feedback cycle between the explainer and explainee highlights two important considerations: 1) interpretability of a black-box model through explanation is dependent on the explainee's mental model \cite{doran2017does, hoffman2018metrics}; and 2) the explainer can generate sequential explanations by leveraging the explainee's mental model \cite{adadi2018peeking, hoffman2018metrics, madumal2018towards}.

We address the problem of irrelevant explanations by modeling the \textit{act of explaining} as a reinforcement learning (RL) task. We propose an RL framework to provide sequential explanations which are relevant to the explainee's current mental model at every iteration, in order to optimize interpretability. Experiments demonstrate enhanced simulatability by selecting explanations using empirically measured observations of the mental model.

\section{Related Work}
The field of \textit{Explainable AI} (XAI) still has many open questions. The \textit{act of explaining} is described as a social interaction between the explainer and the explainee in the form of a dialogue \cite{adadi2018peeking, miller2017explainable}. \citet{miller2017explainable} further proposed that, for an explanation to be effective, significant considerations of the target explainees must be taken into account and explanations should reflect the social behavior of human explanations. Assessment of explanation effectiveness is further challenged from a lack of standard evaluation techniques \cite{adadi2018peeking, doran2017does}. \citet{doshi2017towards} categorized evaluation techniques based on the experimental population and the relevance of the application task. Our experiments use a \textit{human-grounded technique}, where a layperson population is consulted for a generic application task. 

\textbf{Explainers}  
Many explanation methods for black-box models (i.e., \textit{explainers}) have been proposed. In general, explainers are either local or global \cite{adadi2018peeking, doshi2017towards, lipton2016mythos}. While global explanations provide generalized interpretability of model behavior, they may not be as interpretable for specific instances or more complex models \cite{adadi2018peeking}. Explainers are further categorized by their functionality and notation of interpretability, such as saliency maps \cite{montavon2017explaining, binder2016layer, smilkov2017smoothgrad}, example-based \cite{kim2016examples, dhurandhar2018explanations, van2019interpretable, koh2017understanding}, and surrogate models \cite{ribeiro2016should, ribeiro2018anchors}. Given the diversity of explanation types, a criticism is the difficulty of identifying relevant information for a specific explainee \cite{miller2017explainable}. It is impractical for explainees to analyze all possible explanations generated from all explainers to find information relevant to their current mental model. Our framework may select explanations from various explanations and explainers sequentially, in order to provide relevant information to the explainee's current state. 

\textbf{Evaluation Metrics}
Another challenge is the lack of consensus on metrics for interpretability and how effective explanations are to human explainees. Depending on the study, different metrics may be used as proxies for some notion of interpretability \cite{hoffman2018metrics, gilpin2018explaining}. \citet{hoffman2018metrics} identified empirical observations within a conceptual model of the \textit{act of explaining} which may be used as proxies of the explainee's mental model, such as satisfaction in the explanation, trust in the model, human performance on the application task, and human simulatabilty of the model's behavior. Other studies used (or supported the use of) similar metrics as proxies for interpretability \cite{doshi2017towards, lipton2016mythos, gilpin2018explaining, madumal2019explainable, poursabzi2018manipulating, lahav2018interpretable}. We use satisfaction and local simulatability as observations of the explainee's mental model. Any observable and quantifiable evaluation metric, which the user finds reasonable, may also be included into our framework. 

\textbf{Human-in-the-Loop}
Recent studies have included aspects of human experimentation in selecting explanations. \citet{lage2018human} introduced an algorithm for selecting the maximal interpretable model by incorporating human data to establish interpretability priors. \citet{lahav2018interpretable} framed the \textit{act of explaining} as a multi-armed bandit problem for selecting sets of interpretable modules statically (e.g., model attributes and surrogate models) to maximize trust in a domain expert population. By contrast, our framework provides explanations in a sequential fashion, such that each explanation is relevant to the explainee's current, updating mental model.

To our knowledge, this is the first study to measure empirical metrics to observe the explainee's mental model and select explanations. Our experiments, in particular, use a \textit{human-grounded technique} and seek to optimize for simulatability, as an interpretability proxy, within a layperson population.

\section{Framework}
Our goal is to optimize for a specific interpretability proxy, such as simulatability. For each iteration, an agent first observes the current mental model. Following a policy, the agent then selects or generates an explanation, which is optimal for the interpretability proxy. A measure of the interpretability proxy will then be passed to the agent. In practice, the online framework operates sequentially.

The rationale is twofold. Firstly, we hypothesize that providing multiple explanations will increase interpretability for the explainee, as more complementary information may be conveyed. Secondly, we hypothesize that sequential explanations, each provided based on the explainee's updated mental model, will provide more relevant insight to the explainee, and will thus, increase interpretability. This framework ultimately delivers online personalization in providing explanations in a sequential format.

\subsection{Contextual Interpretability as Reinforcement Learning}

\begin{figure}
    \centering
  \includegraphics[width=\columnwidth]{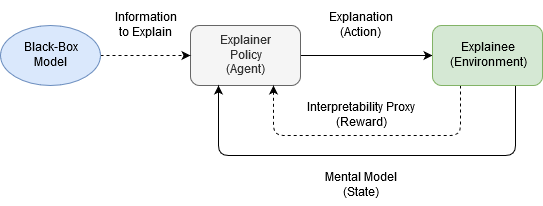}
  \caption{Reinforcement learning representation of the \textit{act of explaining}.}
  \label{fig: rl_framework}
\end{figure}

We model the \textit{act of explaining} of an AI agent to an explainee as a feedback loop (see Figure~\ref{fig: rl_framework}) comprised of three interacting components: the policy $\mu$ (in an RL context, i.e., the \textit{agent}), the explainee who interprets the explanation (i.e., the \textit{environment}), and the black-box model $f(x)$ into which the explanations aim to provide greater insight. The policy provides an explanation $a_{t}$ by observing the context of the mental model $s_{t}$.

We introduce the probability notation for \textit{contextual interpretability}, which includes the mental model as \textit{context} $s_t$. This is the interpretability of a black-box model $f(x)$ from the perspective of a given context and, when context is non-conditional, is closely related to the notation of \citet{lage2018human}. In our framework, we optimize for the interpretability proxy by providing the optimal explanation $a_{t}^{*}$ out of all available explanations $A$:

\begin{equation}
    a_t^* = \arg\max_{a\in A}{p(f(x), a_t\,|\,s_t)},
\end{equation}

The contextual interpretability $p(f(x), a_t\,|\, s_t)$ is the product of the likelihood $p(f(x)\,|\,a_t, s_t)$ and the prior $p(a_t\,|\,s_t)$. Intuitively, $p(f(x)\,|\,a_t, s_t)$ may be defined as the contextual interpretability of the black-box model given an explanation. This is a metric of how interpretable a black-box model is given an explanation and the explainee's mental model. 
This may be rewritten as:

\begin{equation}
    a_t^* = arg\max_{a\in A}{p(f(x)\,|\,a_t, s_t)\cdot p(a_t\,|\,s_t)}
    \label{eq: likelihood_prior}
\end{equation}

\begin{equation}
    a_t^* = arg\max_{a\in A}{p(a_t\,|\,f(x), s_t)\cdot p(f(x)\,|\,s_t)}
    \label{eq: likelihood_evidence}
\end{equation}

where $p(a_t\,|\,f(x), s_t)$ is defined as the contextual interpretability of an explanation given the model (i.e., the posterior). Assuming Markov property, $p(f(x) \, |\, s_t)$ is constant for a given $s_t$ when no explanation is provided. That is, given that the mental model is fully represented by $s_t$, the explainee's understanding of the black-box model given this specified mental model is unchanging. Providing any additional explanations may change $s_t$ and hence, also $p(f(x) \, | \, s_t)$. 

When we optimize for $p(f(x), a_t\,|\,s_t)$, the optimal explanation would maximize the product of the interpretability of the explanation given the mental model and the interpretability of the black-box model given the explanation and the mental model (see Equation~\ref{eq: likelihood_prior}). The interpretability proxy is then the RL expected immediate reward. That is, $r_{t+1}(s_t, a_t) \gets p(f(x), a_t\,|\,s_t)$.

The objective is then to obtain a policy $a_t \leftarrow \mu_{\theta}(s_t)$, such that the expected cumulative reward $Q$ is maximized \cite{watkins1992q}. This optimization may be represented as:

\begin{equation}
    \mu^{*} = \arg\max_{\mu_\theta}{Q^{\mu}(s_t, a_t)}
    \label{eq: optimal_policy}
\end{equation}
\begin{equation}
    Q^{\mu}(s_t, a_t) = E[r_{t+1}(s_t, a_t) + \gamma Q^{\mu}(s_{t+1}, \mu(s_{t+1}))]
    \label{eq: q_function}
\end{equation}

This framework is agnostic to the policies $\mu$, explanations $a_t$, and context state representation $s_t$ used.

\subsection{Implementation}
This section discuss experimental parameters used for our implementation of this framework.

\subsubsection{Explanations}

We define an explanation, $a$, as the output of an explainer, $e(\cdot)$, when one or more data instances $\mathcal{D}$ are inputted (i.e., $a \gets e(\mathcal{D})$). The selection of instances is dependent on the policy and the context.

\begin{figure*}
  \centering
  \includegraphics[width=0.75\textwidth]{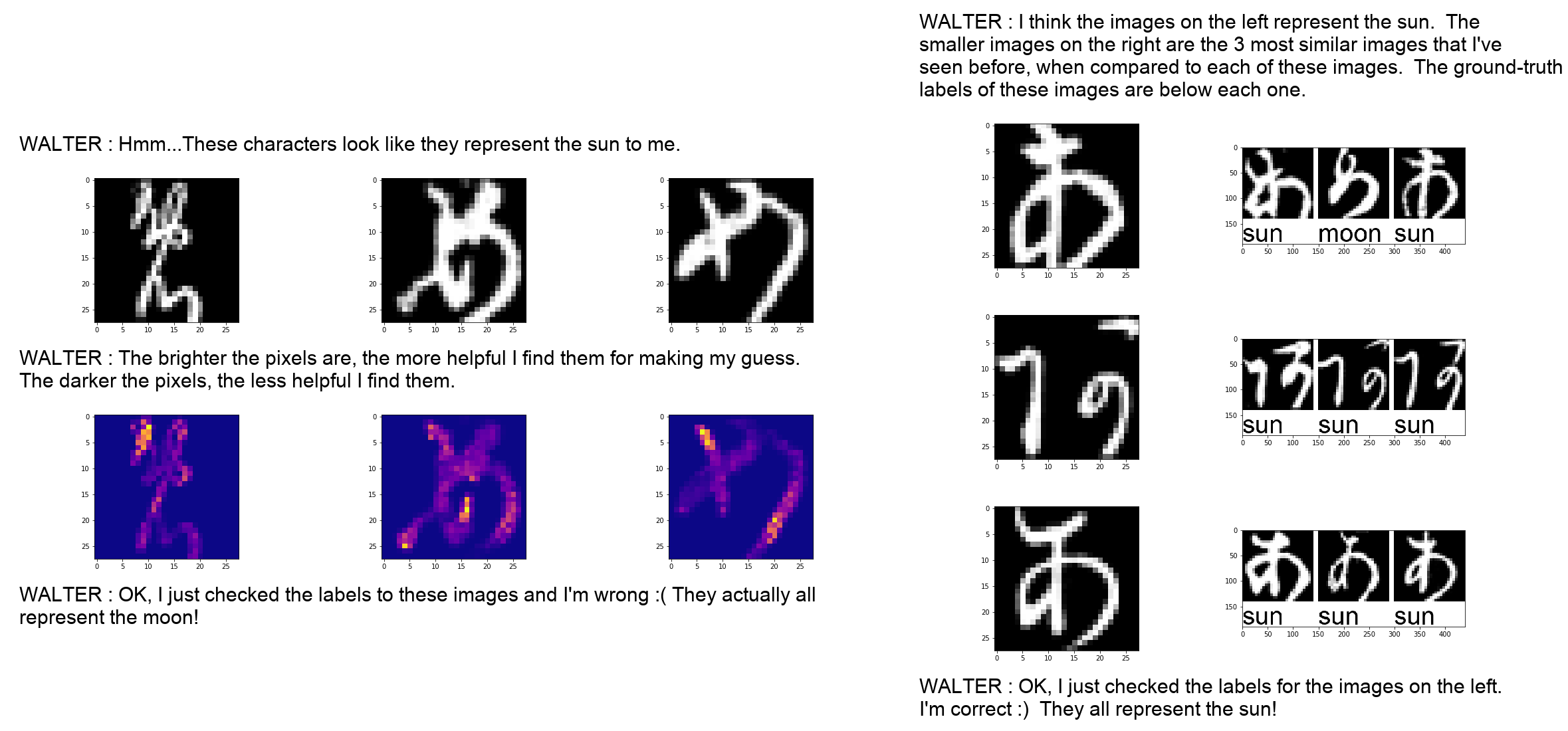}
  \caption{Example explanations presented to explainees. Left - Deep Taylor Decomposition saliency maps for false negatives. Right - Prototypes for true positives. In total, we have 8 different explanations available in our study.}
  \label{fig: heatmap_tp}
\end{figure*}

Our experiments involve eight possible explanations, generated from two local explainers for four sets of data instances (see Figure \ref{fig: heatmap_tp}). The following two explainers were selected due to their popularity within the field of XAI and their demonstrated validity: prototypes \cite{kim2016examples, gurumoorthy2017protodash} and deep Taylor decomposition saliency maps \cite{montavon2017explaining}, implemented using the AIX360 \cite{aix360-sept-2019} and the iNNvestigate \cite{alber2019innvestigate} libraries, respectively. 

Four sets of three instances each represent four classification possibilities of the dataset: true positives (TP), true negatives (TN), false positives (FP), and false negatives (FN). Instance categorization into these classification possibilities allows each explanation to localize on a certain attribute of the feature space, whereas including all classification possibilities avails explanations to represent properties of all test instances. In our implementation, we select the instances which best represent each possibility (e.g., for FP, the three positive instances which have the highest error) for display in the respective explanation. Our pilot studies suggest that explanations consisting of multiple local instances provides greater local representation and human participation than when only consisting of one instance.

\subsubsection{Mental Model and Interpretability Proxy}

To quantify the mental model, we use evaluation metric scores obtained from tasks that the explainee completes every iteration. Because we seek to observe aspects of interpretability from the mental model, it is intuitive that various evaluation metrics for interpretability are used as observations. We use the following, obtained before providing the current explanation, as observations of the mental model (i.e., the context): explainee's satisfaction with the prior explanation \cite{hoffman2018metrics, madumal2019explainable} and explainee's local simulatability of the black-box model for each classification possibility \cite{hoffman2018metrics, doshi2017towards, lipton2016mythos, kim2016examples, gilpin2018explaining, madumal2019explainable, poursabzi2018manipulating, zhu2018explainable}. We use the resultant simulatability, across all classification possibilities (i.e., sum of all local simulatability scores) as our interpretability proxy (i.e., the reward).

\subsubsection{Mental Model-Based Policies}
We implement three experimental policies which select explanations generated by two explainers: 1) \textit{saliency map policy}, which selects the saliency map explanation corresponding to the classification possibility with the lowest local simulatability score in the previous iteration; 2) \textit{prototype policy}, which selects the prototype explanation corresponding to the classification possibility with the lowest local simulatability score in the previous iteration; and, 3) \textit{combined explanations policy}, which selects the explanation where the explainer has the highest mean satisfaction score from the participant and which corresponds to the classification possibility with the lowest local simulatability score in the previous iteration. We compare these policies to random selection baseline policies with the same available explanations.

Our experimental policies simplifies the long-term reward to an immediate reward (i.e., $\gamma=0$ in Equation~\ref{eq: q_function}). By selecting the explanation which greedily increases minimum local simulatability in each iteration, we aim to preserve participation while enhancing immediate simulatability.

\section{Experimental Setup} \label{sec:experimental_setup}

We deploy our framework into an online human experimentation environment. 

\subsection{Dataset and Black-Box Model}
\begin{CJK}{UTF8}{min}

We test our framework with a task which most participants have no experience in, such that we may observe the effectiveness of explanations on the interpretability proxy separate from any prior knowledge. We use a binary classification task on Kuzushiji-49 images \cite{clanuwat2018deep}, which are $28 \times 28$ pixel Japanese Hiragana characters written in cursive form, specifically \textit{a} (あ) and \textit{me} (め) characters, selected for their visual similarities. This dataset was selected due to its classification difficulty for participants (who lack prior knowledge in reading Japanese) compared to other datasets \cite{prabhu2019kannada, deng2009imagenet}, as well as its multi-modal nature, which may assist participants in distinguishing between possible classification possibilities within each character class. Before introducing this task and dataset, participants self-report languages in which they have basic literacy. The task is also described to participants as symbol (i.e., moon-and-sun) classification to reduce association with any prior knowledge of Hiragana characters.
\end{CJK}

A convolutional neural network (CNN) with two sets of layers (i.e., a ReLU-convolutional layer, a batch normalization layer, a max-pooling layer, and a sigmoid-linear layer per set) is used as the black-box classifier. The CNN is trained for 300 epochs with the Adam optimizer \cite{kingma2014adam} and a binary cross-entropy loss function. We achieve an accuracy of 0.85 on a balanced test set after training.

\subsection{Participants}

Participants are recruited through the Amazon Mechanical Turk (MTurk) online crowd-sourcing platform. Participation is open to candidates who possess fluent English literacy, obtained an HIT approval rate greater than 98\%, have over 100 HITs approved, and are located in either Australia, Canada, the United Kingdom, or the United States. 

To qualify for the study, candidates must first complete a pre-assessment, which assesses comprehension of the definitions of explanations and filters out \textit{ineligible} candidates (e.g., random guessing). Candidates are presented examples and descriptions of four types of explanations \cite{montavon2017explaining, kim2016examples, koh2017understanding, ribeiro2016should} on ImageNet images \cite{deng2009imagenet} and are asked comprehension questions. This aims to reduce noise in data quality caused by misunderstanding of the definitions of explanations, as opposed to poor interpretability due to the explanations themselves. In total, 488 participants participated fully or partially across the three experiments. 

\subsection{Experimental Interface and Task Sets}

Participants interact with an iterative survey on Qualtrics. Data is stored using MOOClet \cite{williams2014mooclet}, a back-end engine for running online interactive algorithms. Each of the five experimental iterations first displays an explanation selected by the policy, which is then followed by two task sets, each corresponding to one of the mental model evaluation metrics: satisfaction of the provided explanation and local simulatability of the model's classifaction behavior.

For simulatability task sets, the set of twelve images includes three instances of each classification possibility. This data balancing approach allows the mental model of all $4$ categories to be represented, while preventing success by simply identifying each instance's true label \cite{hase2020evaluating}. We also use a data matching approach \cite{hase2020evaluating} where the image set is consistent for all participants and all iterations. The model predictions and labels of the image set are never revealed to participants. The resultant simulatability score (i.e., the immediate reward) is computed as the sum of all local simulatability scores within an iteration of the task set.

A baseline iteration precedes the five experimental iterations. The baseline does not include the satisfaction task set nor present any explanations, but only a single example image of each class.

\section{Experimental Results}

\begin{figure}
    \captionsetup[subfigure]{position=b}
    \centering
    \subcaptionbox{Saliency maps \label{fig:sub-hm}}{\includegraphics[width=0.75\columnwidth]{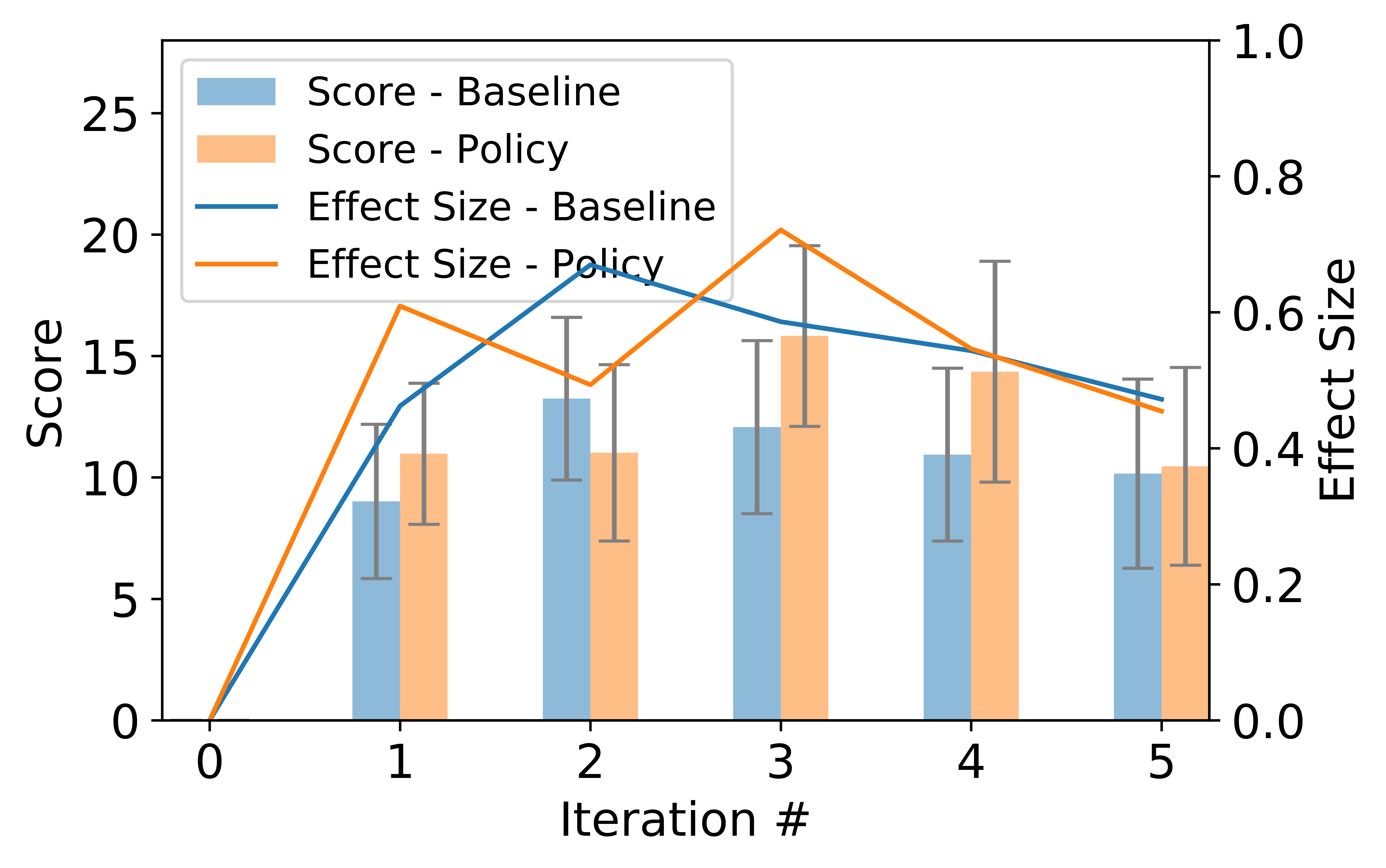}}
    \centering
    \subcaptionbox{Prototypes \label{fig:sub-proto}}{\includegraphics[width=0.75\columnwidth]{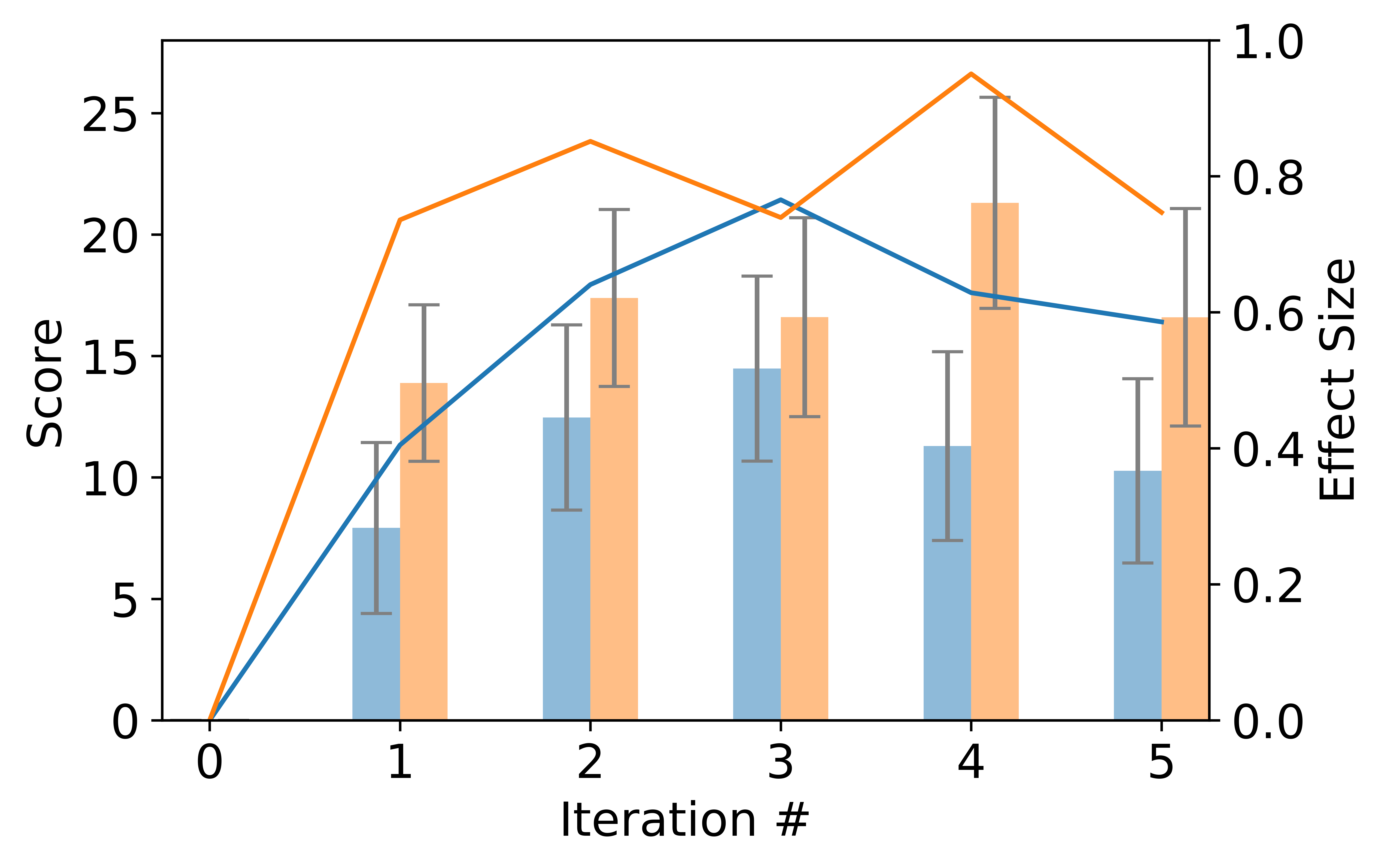}}
    \centering
    \subcaptionbox{Combined explanations \label{fig:sub-multi}}{\includegraphics[width=0.75\columnwidth]{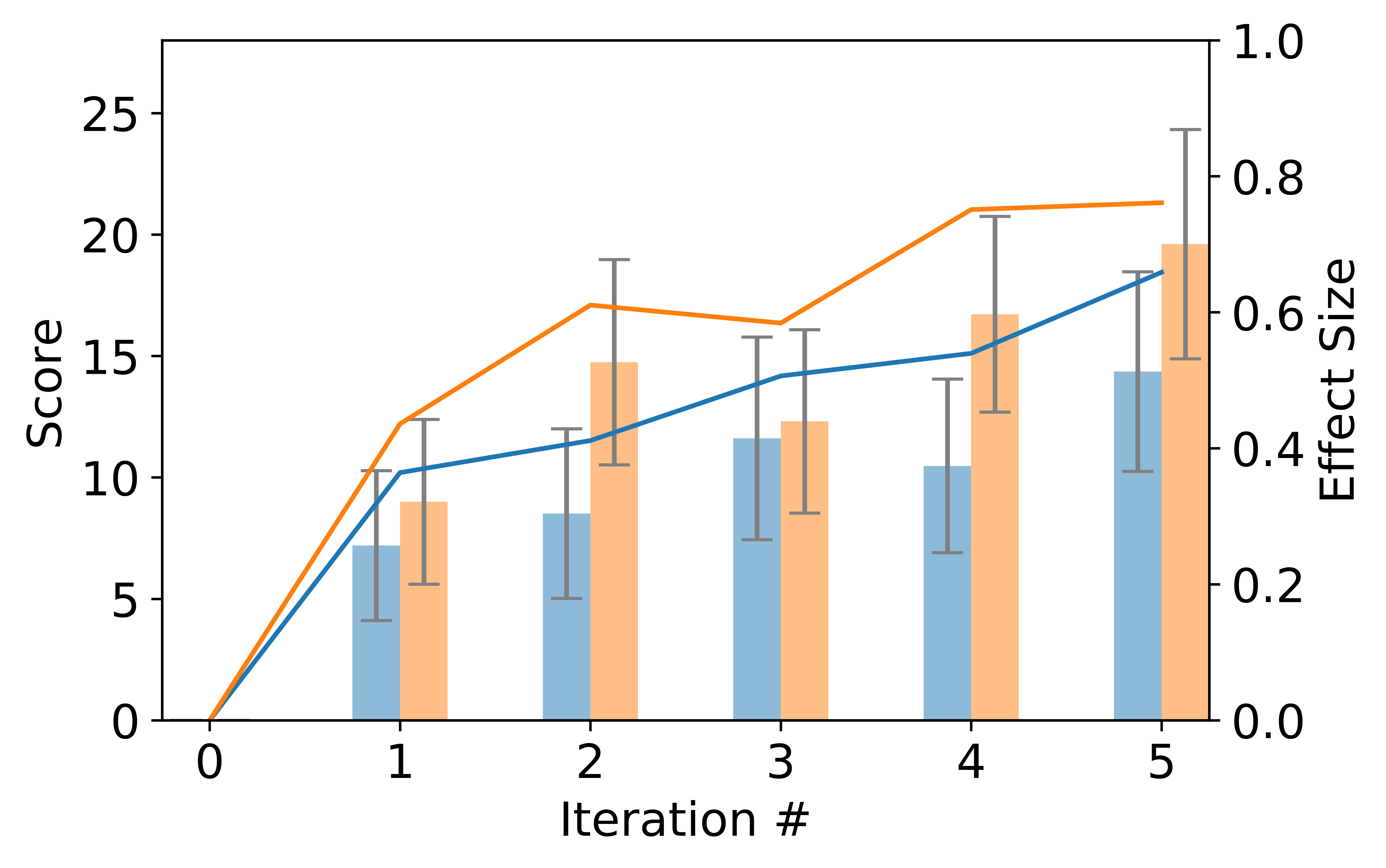}}
    \caption{Mean resultant simulatability scores (i.e., the reward) and effect sizes, across all participants, for every iteration. Resultant simulatability scores for each participant are relative to their baseline score. Likewise, effect sizes for each iteration are relative to the mean resultant simulatability score of the baseline iteration. Error bars represent standard error.}
    \label{fig:progress}
    \vskip -0.2in
\end{figure}

\textbf{Different policy-explanation combinations may result in varying trajectories.} From the three experiments, we observe initial increases in the simulatability score relative to the baseline iteration in all trajectories (see Figure \ref{fig:progress}). The combined explanations experiment shows an upward trend, whereas the other two experiments show gradual plateauing of the simulatability score. The lack of difference between the trajectories in the saliency map experiment may indicate a lack of effectiveness of the policy for deep Taylor decomposition saliency map explanations, whereas for the other two experiments, the greater simulatability scores observed in the mental model-based policy trajectories suggest policy-explanation compatibility. The continual upward trend in the combined explanations experiment may also indicate that multiple explanation types may provide complementary relevant information to the explainee, or may select explanations which match the explainee's individual preferences, compared to the other two experiments where each uses a single explainer.

\textbf{Mental model-based policies may assist in increasing simulatability over time.} While the saliency map approach shows similar simulatability scores for both the mental model policy and the baseline, the prototypes and combined explanations approaches show more noticeable differences between the mental model policy and the baseline (see Figure \ref{fig:progress}). In these two experiments, the mean simulatability scores of the mental model-based policies are consistently greater than those of the baselines. This difference is reflected in their effect sizes, computed with Cohen's $d$ \cite{cohen1988lawrence}, relative to their initial iterations. The greater effect sizes of mental model-based policies suggest the effectiveness of these policies. Additionally, medium to large effects ($d > 0.5$) in both trajectories suggest the effectiveness of the explanations themselves. High standard errors, however, indicate high variance, possibly due to non-conformity in the effect of explanations on human participants, as well as other experimental factors which could not be monitored through online experimentation (e.g., fatigue, attention).

The gradual plateau of simulatability scores in the saliency map and prototype experiments may suggest either a lack of new relevant information available from additional explanations or inherent degradation in simulatability over time, which are counteracted by information provided by additional explanations. Intuitively, an increase in interpretability after providing the initial explanation is expected, as it will provide new information regarding the feature space to the explainee who has no prior knowledge. Following explanations may not provide as much new information, depending on how similar they are to previous explanations. Additionally, factors that may degrade interpretability over time may include information overload, shift in attention to different local regions in the feature space, or participant fatigue. In theory, such factors may be used as mental model observations to provide better representations of the context. We do not observe plateauing of simulatability scores in the combined explanations experiments, suggesting different explanation types may provide less similar, yet complementary information regarding the feature space.

\section{Conclusion}

We propose a sequential RL framework for explaining the behavior of a black-box model by an AI agent to a human explainee. This framework optimizes for a specific interpretability proxy which is measured from the explainee. In every iteration, metrics of the explainee's mental model are observed by an explainer policy to select or generate an explanation which will optimize for the interpretability proxy. We deployed online human-interaction experiments of this framework on Amazon MTurk. We compared three experimental policies in selecting explanations generated by prototype \cite{kim2016examples, gurumoorthy2017protodash} and deep Taylor decomposition saliency map \cite{montavon2017explaining} explainers, relative to random selection baselines.

Our results suggest potential effectiveness of providing sequential explanations using policies which observe the explainee's current, updating mental model. Depending on the explanation type(s), mental model-based policies may lead to greater simulatability than random selection. 

Our experiments on MTurk, however, lack participant monitoring, which may lead to greater variance in our data, as observed in our simulatability measures. While our experiments are limited to pre-defined policies and two explanation types, future research may include training policies using multi-armed bandit or deep reinforcement learning techniques, which may require larger amounts of behavioral data. Additionally, we focus on the simulatability of a specific classification task. While our results may be relevant to similar tasks and datasets, further experimentation is necessary to guarantee the framework's applicability to more diverse tasks and datasets.   

This framework provides a model for mapping the \textit{act of explaining} into a dialogue-like process between an explainer AI agent and the human explainee. When feedback regarding the explainee's mental model is provided to the explainer, our framework may be used to provide explanations which are maximally relevant and useful to the explainee. This will increase the interpretability of black-box models, while limiting irrelevant information. 

\section*{Acknowledgments}
We thank Sam Maldonado for setting up the MOOClet engine back-end server for data collection. This work was supported by Electronics and Telecommunications Research Institute (ETRI) grant funded by the Korean government [20ZS1100, Core Technology Research for Self-Improving Integrated Artificial Intelligence System] and the Canadian Institute for Advanced Research (CIFAR).

\bibliography{references}
\bibliographystyle{icml2020}



\end{document}